\documentclass[conference]{IEEEtran}
\IEEEoverridecommandlockouts
\usepackage{cite}
\usepackage{amsmath,amssymb,amsfonts}
\usepackage{subfigure,bm,balance}
\usepackage{algorithmic}
\usepackage{graphicx}
\usepackage{textcomp}
\usepackage{xcolor}
\usepackage{multirow}
\usepackage{multicol}
\def\BibTeX{{\rm B\kern-.05em{\sc i\kern-.025em b}\kern-.08em
    T\kern-.1667em\lower.7ex\hbox{E}\kern-.125emX}}
\begin{document}
\setlength{\abovedisplayskip}{3pt}
\setlength{\belowdisplayskip}{3pt}

\title{OPAM: Online Purchasing-behavior Analysis using Machine learning \\
}

\author{\IEEEauthorblockN{Sohini Roychowdhury}
\IEEEauthorblockA{\textit{Director, ML Curriculum} \\
\textit{FourthBrain.ai, CA-95050, USA}\\
Affiliate, Univ. of Washington, roych@uw.edu}
\vspace{-0.5cm}
\and
\IEEEauthorblockN{Ebrahim Alareqi}
\IEEEauthorblockA{\textit{Data Scientist, Product Labs} \\
\textit{Volvo Cars Technology, CA}\\
ealareqi@volvocars.com}
\vspace{-0.5cm}
\and

\IEEEauthorblockN{ Wenxi Li}
\IEEEauthorblockA{\textit{Graduate Student Researcher, Data-X Lab} \\
\textit{University of California, Berkeley}\\
liwenxi@berkeley.edu}
\vspace{-0.5cm}
}

\maketitle

\begin{abstract}
Customer purchasing behavior analysis plays a key role in developing insightful communication strategies between online vendors and their customers.
To support the recent increase in online shopping trends, in this work, we present a customer purchasing behavior analysis system using supervised, unsupervised and semi-supervised learning methods. The proposed system analyzes session and user-journey level purchasing behaviors to identify customer categories/clusters that can be useful for targeted consumer insights at scale. We observe higher sensitivity to the design of online shopping portals for session-level purchasing prediction with accuracy/recall in range 91-98\%/73-99\%, respectively. The user-journey level analysis demonstrates five unique user clusters, wherein \textit{New Shoppers} are most predictable and \textit{Impulsive Shoppers} are most unique with low viewing and high carting behaviors for purchases. Further, cluster transformation metrics and partial label learning demonstrates the robustness of each user cluster to new/unlabelled events. Thus, customer clusters can aid strategic targeted nudge models.
\end{abstract}


\section{Introduction}
Recent times have witnessed a significant increase in online shopping activities, that in turn has necessitated development of personalized recommendation systems for a seamless shopping experience \cite{onlineref}. As shoppers continue to interact with products online, purchasing behavior patterns can be mined from session-level \cite{onlineref2} and user-product journey level interactions for targeted communications and personalized shopping strategies for vendors and customers, respectively. For example, accurate purchasing predictions from user groups can ensure adequate product inventory based on user engagement levels and it can lead to directed nudge models beneficial to a specific user group. 

The primary bottleneck for designing such a predictive system is the disparity in virtual shopping experiences across products and platforms due to variations in product cost, delivery wait times and ease of platform usage \cite{data1} \cite{data2}. In this work, we present a novel online shopping behavior analysis system that utilizes supervised, unsupervised and semi-supervised learning models to identify and analyze \textit{purchasing behavior} clusters/categories for accurate inventory and strategic marketing campaign design. We demonstrate the scalability of the proposed system across product cost, category, wait time and data size constraints by analyzing purchasing behaviors for cosmetics \cite{data1} and electronics data sets \cite{data2} separately. The proposed system provides detailed insights into customer-product interactions at a session and journey level by assessing three properties: 1) importance of shopping platform and product-level features for purchase predictions, 2) formation of user-behavior clusters and the effort required to transform users across clusters, 3) the robustness of sample neighborhoods toward unlabelled/partially labelled data sets representing high volumes of ongoing user journeys.

\begin{figure}[ht]
\centering
\includegraphics[width=0.52\textwidth]{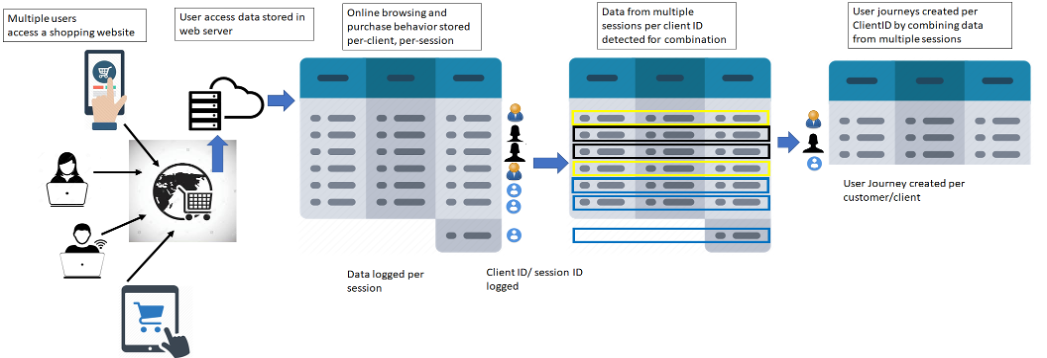}
\caption{Example of data flow to create user-journeys from session-level data.}
\label{fig}
\vspace{-0.3cm}
\end{figure}

Each time customers login to an online shopping website, they accept cookies to establish the session. The \{session-ID, client-ID\} combination can then be used to uniquely log information regarding product-level browsing, additions to cart, removals from cart and purchase etc. The session-level data can then be consolidated to create user-journeys, as shown in Fig. \ref{fig}, to predictively analyze the purchasing propensity per user-type for each product-type interaction.

 Existing works so far in \cite{onlineref} \cite{onlineref2} have analyzed products with demand and price variations such as cosmetics vs. electronic items at session-levels differently. The work in \cite{clicknn} applied the time-stamps of clicks in a clicking stream per session to model the buying patterns and then predicted purchasing decisions using bidirectional LSTM models. Further, \cite{clicknn} demonstrated that the accuracy from click stream sequences and LSTM models was comparable to that using engineered features and classification models. In this work, we extend the analysis of predictive models at session-level and user-journey level for purchase events in terms of sensitivity to the online shopping portal and product-level features, respectively.

Apart from session-level assessments, the work in \cite{repeatbuy} shows that there is a need to predict repeat customers and their tendency to return and finish their orders. This motivates our analysis of user-journey level interactions. Here, we analyze session-level and user-journey level features to categorize user-interaction clusters. The knowledge of user clusters can thus lead to improved predictive modeling for purchasing events per cluster to accurately gauge customer-specific demands. 

Additionally, probabilistic label propagation using k-NN, SVM and other methods, has been used so far to assess the sensitivity of training data to controlled false labels in existing works \cite{PLL1}. The major difference between the datasets in \cite{PLL1} \cite{PLL2} and this work is that false labels are not naturally occurring for user-journeys. Instead, we use semi-supervised methods to analyze predictive robustness for each user cluster, i.e., how accurately the purchasing behavior can be predicted in case of lost or missing labels.

\begin{figure}[ht]
     \centering
    \includegraphics[height=1.4in, width=3.0in]{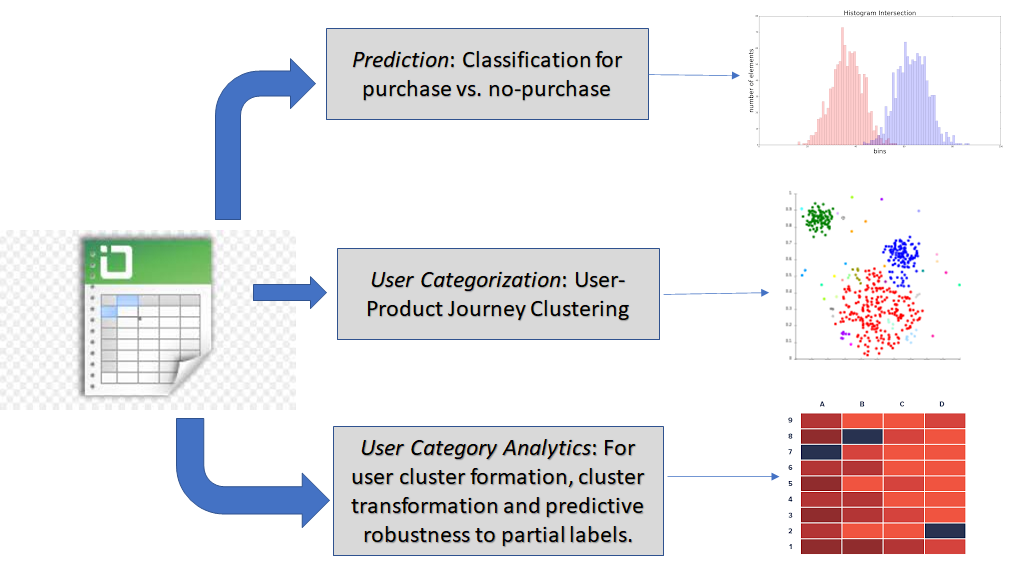}
        \caption{Proposed system for analysis of user-behavior categories/clusters. Step 1: To find optimal classifiers for predicting sessions and user-journeys leading to a purchase. Step 2: Identification of user clusters. Step 3: Analysis of user clusters for formation, transformation and robustness to unlabelled data.}
    \label{model}
    \vspace{-0.4cm}
\end{figure}

The proposed system shown in Fig. \ref{model} makes the following three major contributions. First, we analyze the sensitivity of the online shopping portal and product-level features for session-level and user-journey level classification, respectively, for purchase prediction. 
Second, we apply unsupervised learning to identify unique user-behavior clusters/categories. We observe that predictability of a purchase event per cluster varies significantly. Third, we perform user-cluster analytics that includes cluster formation, effort required for inter-cluster transformation and predictive robustness per cluster using k-NN based label propagation. The Python code for our system is provided for benchmarking and system extendability \footnote[1]{https://github.com/sohiniroych/Volvo-DataX}.

\section{Materials and Methods}
Descriptions of the data sets under analysis and the methods used to analyze the session and user-journey level data sets are described below.

\subsection{Data and Output Metrics}
We analyze two public datasets acquired from Kaggle for a comprehensive experiment series, including eCommerce Events History in Cosmetics Shop \cite{data1} (size 2 GB), as well as eCommerce behavior data from multi category store \cite{data2} (size 15 GB). For the second dataset, electronics data is retained to keep the study market specific. Both datasets have the same columns: user ID, time stamp of the event, product metadata, i.e., \{product category, brand, price\}, user session ID and event type. For the cosmetics data set, the event types are \{cart, view, remove\_from\_cart, purchase\}. However, for the electronics data set, event types are \{cart, view, purchase\}. This variation in data event types aids investigations towards the predictability of a purchasing event based on differing online shopping platforms.

For our analysis, we perform data aggregation to extract features corresponding to each user-product interaction at a session-level and at user-journey levels (denoted by $\{X, X^s\}$, respectively). If the session or user-journey contained a purchasing event, the label $\{Y,Y^s\}$, is set to 1, else if the session or journey resulted in product viewing, carting and or removing from cart, then the label is set to 0, respectively.

The session-level data and user journey-level data is subjected to feature ranking followed by data modeling using an AutoML library\cite{TPOT} to identify the \textit{best data model} that can predictively identify purchasing sessions and journeys from the non-purchasing ones. Given, that the classification models result in true positive ($tp$), true negative ($tn$), fale positive ($fp$) and false negative ($fn$) samples, the output metrics are $accuracy=\frac{tp+tn}{tp+tn+fp+fn}, precision=\frac{tp}{tp+fp}, recall=\frac{tp}{tp+fn}, F1=\frac{2tp}{2tp+fp+fn}$.

The data sets under analysis here demonstrate high data imbalance with  the ratio of non-purchase to purchase records being nearly 7:1 in the cosmetics dataset and 35:1 in the electronics dataset when aggregated at a user-journey level. At session-level aggregation, this ratio is 28:1 and 16:1 in the cosmetics and electronics datasets, respectively.
Thus, the metrics that need to be maximized for \textit{best data model} are $recall, F1$ metrics that place a higher weightage on accurate detection of \textit{purchase} records over the \textit{non-purchase} records. 
\subsection{Session-Level Classification}
There are two specific instances where session-level predictive modeling is more beneficial than user-journey level. First, if a user logs in through multiple devices or skips the login process altogether, multiple session records get created wherein the history cannot be retained. In such cases session-level predictions for a purchasing event can guide marketing nudge models. Second, sequence models such as LSTM are capable of learning local and global contextual patterns for purchasing behaviors. Thus, data processing on a session-level can prove to be more storage efficient than user-journeys. 

As a first step, hand-engineered features are subjected to feature ranking using Random forest and Fisher scoring \cite{ranking}. The session-level features selected for cosmetics dataset are: total number of events (viewing, carting removal etc.), number of brands in cart, number of products in cart, number of times \textit{carting} occurs, number of times \textit{removal from cart} occurs, number of \textit{viewing} events, number of brands viewed, and number of products viewed.
Similarly, the features selected for electronics dataset are: average price of the products in cart, number of brands in cart, number of categories in cart, number of products in cart, number of times \textit{carting} occurs, total price in cart, number of events in a session, total interaction time, and number of brands viewed. Next, based on session-level features, LSTM-based models are used to classify a purchase at session-level.

\subsection{User-Journey Level Classification}
\begin{figure}[ht]
    \centering
	\subfigure[Cosmetics dataset]
	{\includegraphics[width=0.22\textwidth]{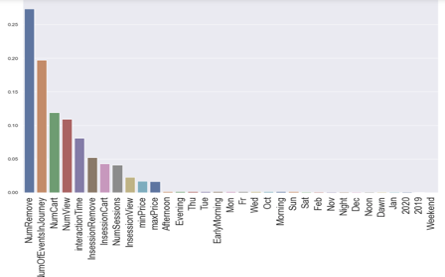}}
	\subfigure[Electronics dataset]
	{\includegraphics[width=0.22\textwidth]{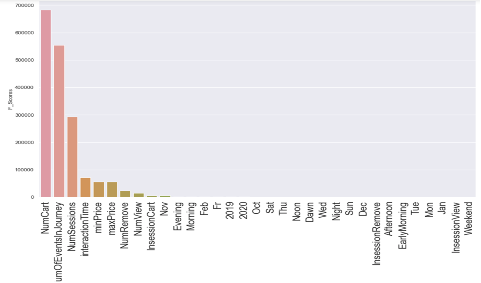}}
	\caption{Feature Ranking using (a) Random Forest and (b) Fisher Score for Journey Features on Cosmetics data set, respectively.}\label{rank}
       \vspace{-0.4cm}
\end{figure}

Similar to the session-level data, the user-journey level data is subjected to feature ranking \cite{ranking}. Exploratory data analysis and feature ranking demonstrates that there is little to no variance in the distribution of purchase journeys vs. non-purchase journeys when measured against date-time attributes. An example of feature ranking at user-journey level is shown in Fig. \ref{rank}, where we observe that features like number of events in user-journey, total interaction time, number of sessions, number of carts, views and removals have significantly higher weightage than features like date, time or month of purchase.

Based on this analysis, we select the top ranked 11 features with significant weights that range from total interaction time, number of events, total carting and viewing time to max and min price range. Each feature is then scaled in the range [0,1] for further data modeling and analysis.
\subsection{Clustering of User-journey data}\label{clus}
We perform unsupervised clustering on the user-journey data, where, the optimal feature sets extracted above are converted to the t-SNE plane. First, we apply k-means clustering to the converted data samples followed by the Elbow method (using the Yellowbrick Library in Python) to find the optimal number of user-journey clusters (Q) that minimizes overall distortion score per cluster. For both cosmetics and electronics data sets, the optimal number of clusters identified are $K=5$, as shown in Fig \ref{yellow} below. 

\begin{figure}[ht]
     \centering
    \includegraphics[height=1.8in, width=3.0in]{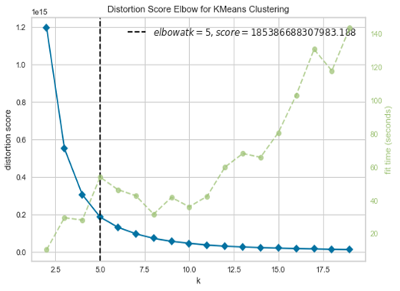}
        \caption{Elbow method to find $K=5$ clusters for the Electronics dataset. Similar curve is observed for the Cosmetics dataset.}
    \label{yellow}
    \vspace{-0.3cm}
\end{figure}

\subsection{User-Journey Cluster Analytics}\label{metrics}
The user journey data is \{$X,Y$\}, where the journey-level features, $X \in \mathcal{R}^{[nXd]}$, and the journey's purchasing outcome is represented by   $[Y=(0,1) \in \mathcal{R}^{[nX1]}, \forall i=1:n] $. Here, $n$ represents the number of samples and $d$ represents the selected feature dimensions. Next, the user-journey cluster ID for each sample can be treated as a feature vector representative of the sample neighborhoods as, $Q \in \mathcal{R}^{[nX1]}$. It is noteworthy that the distribution of sample clusters is non-linearly dependent on the user-journeys. Hence, combining user journeys with cluster information should intuitively reduce model fitting error.

To analyze the user-clusters, we perform the following three sets of experiments using the samples per cluster ($X_q, Y_q),  \forall q=[0,1,..K-1]$) defined in \eqref{sub}. Here, $n_q$ represents the number of samples in cluster $q$.
\begin{align}\label{sub}
(X_q,Y_q) \subset (X,Y), s.t. (X_q,Y_q)=\{(x_q(i'),y_q(i'))\},\\ \nonumber
\text{where}, Q(i')=q,  i'=[1:n_q]. \\ \nonumber
\end{align}
\subsubsection{Cluster Formation Analysis}
We analyze the formation of the user-clusters in terms of Calinski Harabasz ($CH$) score defined in \eqref{CH} and Silhouette Scores ($SS$) defined in \eqref{SS} based on prior works in \cite{cluster}. High $CH$ scores are representative of maximum between cluster separations ($B_q$) and minimum within cluster distances ($W_q$) defined in (3). Here, $\bar{x}$  and $\bar{x_q}$ imply total sample mean and sample mean per cluster, respectively.
\begin{align}\label{CH}
CH=\frac{tr(B_q)}{tr(W_q)} \frac{(n-K)}{(K-1)}, \\ 
\text{where, } W_q=\sum_{q=0}^{K-1}\sum_{i'=1}^{n_q}(x_q(i')-\bar{x_q})(x_q(i')-\bar{x_q})^T\\ \nonumber
 B_q=\sum_{q=0}^{K-1}n_q(\bar{x_q}-\bar{x})(\bar{x_q}-\bar{x})^T\\ \nonumber
\end{align}
Consistency in the $SS$ metric across cluster definitions is indicative of large separating distances between samples of different clusters (represented by $b$) over samples of the same cluster (represented by $a$) in (5). Here, $x(i') \in \mathcal{R}^{[1Xd]}$.
\begin{align}\label{SS}
SS=\frac{(b-a)}{\max(a,b)}, \text{where, }\\ \nonumber
a=\frac{1}{n_q}\sum_{x(i') \in X_q,}\sum_{x(j) \in X_q, i'\neq j}(x(i')-x(j))(x(i')-x(j))^T\\
b=\frac{1}{n_q}\sum_{x(i') \in X_q,}\sum_{x(j) \notin X_q}(x(i')-x(j))(x(i')-x(j))^T\\ \nonumber
\end{align}

\subsubsection{Cluster Transformation Analysis}
Next, we analyze the effort required to transform a cluster distribution to another cluster in terms of the EarthMover-Distance (EMD, or Wasserstein-1 distance) \cite{EMD}. EMD is an asymmetric measure and is computed between each combination pairs of clusters, between cluster $q$ and $q'$ using representative histogram distributions in \eqref{EMD}. Here, scaled samples ($X_q$) are subjected to binning operation performed for each feature, such that $P_q(h)$ represents the $h$ th bin for the histogram created for all samples across all features in cluster $q$. Here, we limit the total number of bins to $H=10^6$. Also, in \eqref{EMD}, $F_q(h)$ represents the quantile function at bin value $h, \forall  h=[0,10^{-6}, 2*10^{-6}...1]$.
\begin{align}\label{EMD}
EMD(P_q,P_{q'})=\sum_{h=1}^{H}|F_q(h)-F_q'(h)|\Delta h\\
\text{where, } F_q(h)=\sum_{l=1}^h P_q(l).\\ \nonumber
\end{align}
High values for EMD imply large effort required to transform samples across the clusters and vice versa.
\subsubsection{Predictive Robustness using Partial Label Learning}
Finally, we analyze the predictive robustness of each cluster to missing labels or ongoing user-journeys using partial label learning (PLL) in the semi-supervised setting as in \cite{PLL1}. For this analysis, we randomly select user-journey samples per cluster $X_q$ and drop their labels (assign $y=-1$). Next, we apply k-NN based label propagation (here, k-varied as odd numbers in range (1:15)) with $\alpha=0.1$, which is representative of low modification rate to propagated labels in a transductive setting \cite{PLL2}. Performing 5-fold cross validation, we determine that $k=3$ results in least prediction error. Next, to analyze predictive robustness per cluster, we sub-sample instances by varying the proportion of dropped labels $p$ in the range [0.1-0.9]. Next, we apply label propagation for each sub-sampled instance followed by plotting the $accuracy$ and $F1$ for classification of the samples with dropped labels. This process is repeated 50 times per drop-proportion $p$ and the averaged $accuracy$ and $F1$ scores are plotted. Here, the goal is to assess the consistency in trends of $accuracy$ and $F1$ across clusters to identify which clusters are robust to missing labels.
%%%%%%%%%%%%%%%%%%% Here%%%%%%%%%%%%%%%%%
\section{Experiments and Results}
Our primary goal is to assess the predictability of a purchase event at a session and user-journey level in terms of features and sample neighborhood. Thus, we perform three sets of experiments corresponding to the three modules in the proposed system. First, we implement sequence models to predict purchasing events on a session-level. The probabilistic outcome per-session indicates if the next session is likely to result in a purchase or not. Second, we implement classification models on a user-journey level using TPOT library \cite{TPOT}. Third, we analyze the user journey clusters to infer customer insights that are necessary to inform marketing and customer retention strategies.

\subsection{Session-Level Classification}
For the session-level data, we implement LSTM models with differing layer and neuron structure combinations to identify the best LSTM model structure for both data sets. At a session-level, the percentage of sessions that end in purchases are 9.22$\%$ and 10.94$\%$ for the cosmetics and electronics data sets, respectively. Thus, in the absence of a trained LSTM model, if all sessions were assigned the major non-purchase class, we would still achieve 90.78$\%$ and 89.06$\%$ baseline accuracy for the cosmetics and electronics data sets, respectively. To counteract the class imbalance, LSTM models are trained on balanced data generated by over sampling the minor class distribution \cite{ranking}. Here, a variety of LSTM network structures are analyzed, with 1-3 layers of bidirectional LSTM layers and 10-40 neurons per layer. 
	
The network structure resulting in the highest $recall$ is presented in Table \ref{res2}. We empirically determine that a single layer of bidirectional LSTM with 40 neurons is the best model for the cosmetics and electronics data sets, respectively. From Table \ref{res2}, we observe that prediction of a purchase event is more accurate (high $recall$, $F1$) for the cosmetics data set than for the electronics dataset. One reason for the high $fp$ rate in the electronics data set is that \textit{view} events take up almost 95$\%$ of all events, and there is no option to \textit{remove$\_$from$\_$cart}. Thus, a viewing session most often gets falsely predicted to end up in a purchase event. Additionally, to baseline the session-level prediction of a purchasing event, we create a sequence model from the sequence of product-level interactions per session and the time spent per interaction from \cite{clicknn}. To create the baseline sequence model, the session-level events are categorized as: {1=view, 2=cart, 3=remove from cart, 4=purchase}. So the input ($X^s$) is a sequence of maximum 100 such events without the purchasing event e.g. $\{$1,2,1,1,1,1,3,1,,......$\}$, and the time spent on each event, while the output ($Y^s$) is binary representing a purchase event occured or not. 
\begin{table}[ht]
\centering
\caption{Session-level classification. Best Values in bold.}
\scalebox{0.85}
    {
 \begin{tabular}{|c|c|c|c|c|}
 \hline
Model&$recall$&$accuracy$&$precision$&$F1$\\ \hline
\multicolumn{5}{|c|}{Cosmetics Dataset}\\ \hline
Proposed (1 layer, 40 neurons)&\bf{0.9999}&{\bf0.9796}&0.7733&\bf{0.8722}\\ \hline
Sequence Model&0.1402&0.9206&\bf{0.9960}&0.2458\\ \hline
\multicolumn{5}{|c|}{Electronics Dataset}\\ \hline
Proposed (1 layer, 10 neurons)&\bf{0.7344}&{\bf0.9162}&0.4392&{\bf0.5497}\\ \hline
Sequence Model&0.2581&0.9017&\bf{0.6233}&0.3650\\ \hline
 \end{tabular}
 }
 \label{res2}
 \vspace{-0.3cm}
\end{table}
We observe that when compared to proposed Bi-LSTM on session-level features, the sequence models fail to capture product price and brand-related information, that are key factors impacting purchasing decisions. This results in significantly higher session-level $recall$ for the proposed model over the baseline sequence models on both data sets. Lower $precision$ for the proposed model may lead to some over-stocking inventory per session that would not impact the online shopping experience negatively.
\subsection{User-journey level classification}\label{classify}
Both the cosmetics and electronics are benchmarked using AutoML TPOT library (v 0.11.6.post3) \cite{TPOT}. Each data set is sub-sampled by stratified sampling based on the cluster IDs into samples of 25,000 each with the selected set of features after feature ranking. These sub-samples are subjected to the AutoML to identify the \textit{best data model} with optimal hyperparameters. The  \textit{best data model} is then applied to the complete data set in batches of 1-2M samples each, generated by cluster-based stratified sampling. The average classification performances after 25 such runs of stratified sampling for 70/30 train-test split are shown in Table \ref{journeyresult}. For the cosmetics data set, the XGboost Classifier is found to be the \textit{best model} with the following parameters: (learning rate=0.001, max depth=9, min child weight=7, n\_estimators=100, n\_jobs=1). For electronics data, the  \textit{best model} is Decision Tree Classifier with (max depth=10, and minimum samples leaf=3, min samples split=2). The average classification performances of the \textit{best data models} in comparison with a baseline k-nearest neighbor (k-NN) classifier with $k=3$ is shown in Table \ref{journeyresult}.
\begin{table}[ht]
\centering
\caption{User-Journey Classification Performance. Best values in bold.}
 \begin{tabular}{|c|c|c|c|c|}
 \hline
Model&$accuracy$&$precision$&$recall$&$F1$\\ \hline
\multicolumn{5}{|c|}{Cosmetics Dataset}\\ \hline
XGBoost&\bf{0.9377}&{\bf0.8949}&{\bf0.5487}&{\bf 0.6803}\\ \hline
kNN&0.8861&0.5336&0.4578&0.4928\\ \hline
\multicolumn{5}{|c|}{Electronics Dataset}\\ \hline
DecisionTree&\bf{0.9997}&0.9931&\bf{0.9883}&{\bf0.9907}\\ \hline
kNN&0.9994&{\bf0.9933}&0.97061&0.9818\\ \hline
 \end{tabular}
 \label{journeyresult}
\end{table}
Here, we observe that although the electronics data set has higher purchase to non-purchase imbalance, it has better classification performance for purchase events than the cosmetics data set. This is primarily due to the fact that product-level feature variations are significantly lesser among cosmetics samples. Most cosmetics have a narrow price range of (5\$-11\$) while electronics have a wider range of (290\$-402\$). Additionally, the \textit{brand-specific} distinction in electronics is significantly higher than for the cosmetics data set. Thus, product-level features have a significant impact on user-journey level prediction of purchase events.
\subsection{User-journey Cluster Analysis}
\begin{figure}[ht]
    \centering
	\subfigure[Cosmetics Dataset]
	{\includegraphics[width=0.22\textwidth]{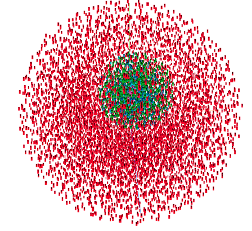}}
	\subfigure[Electronics Dataset]
	{\includegraphics[width=0.22\textwidth]{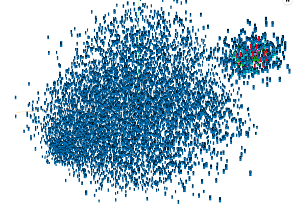}}
	\caption{User-journey based clusters in t-SNE space.}\label{visual}
       \vspace{-0.4cm}
\end{figure}

From the data, we know that at user-journey level the complete cosmetics and electronics data sets have purchase ratios around 12\% and 1\%, respectively. Also, data clustering provided 5 aggregated user clusters for each data set in Section \ref{clus}. Sample visualization of these user clusters is shown in Fig. \ref{visual}.

\begin{figure*}[ht!]
    \centering
	\subfigure[Cosmetics Data set]
	{\includegraphics[width=0.83\textwidth]{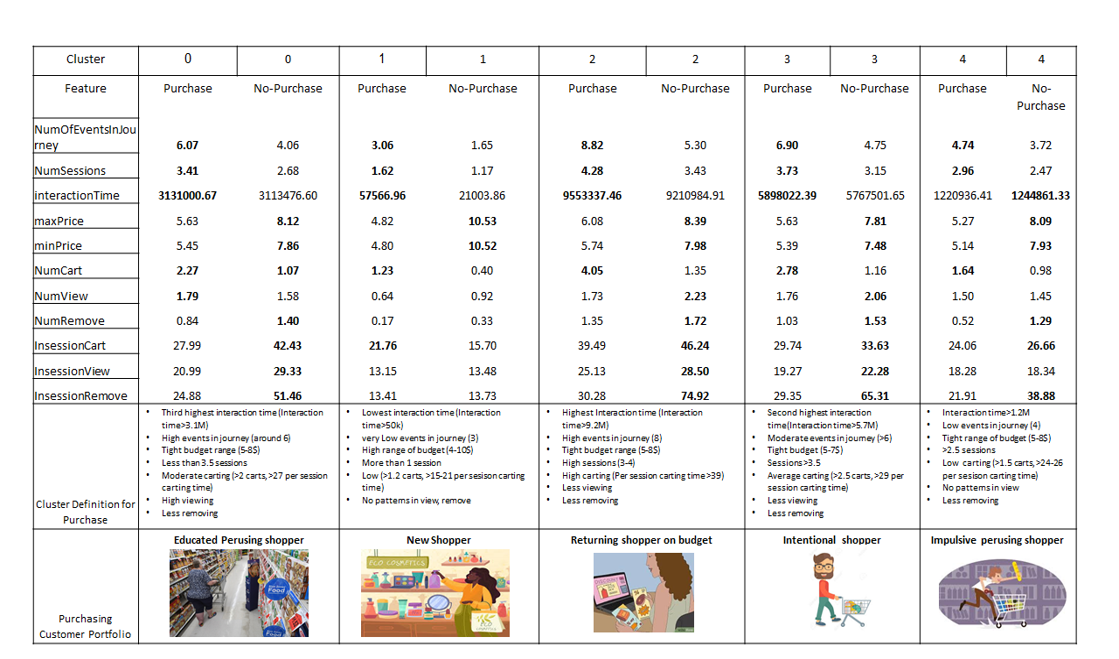}}
	\subfigure[Electronics Data set]
	{\includegraphics[width=0.83\textwidth]{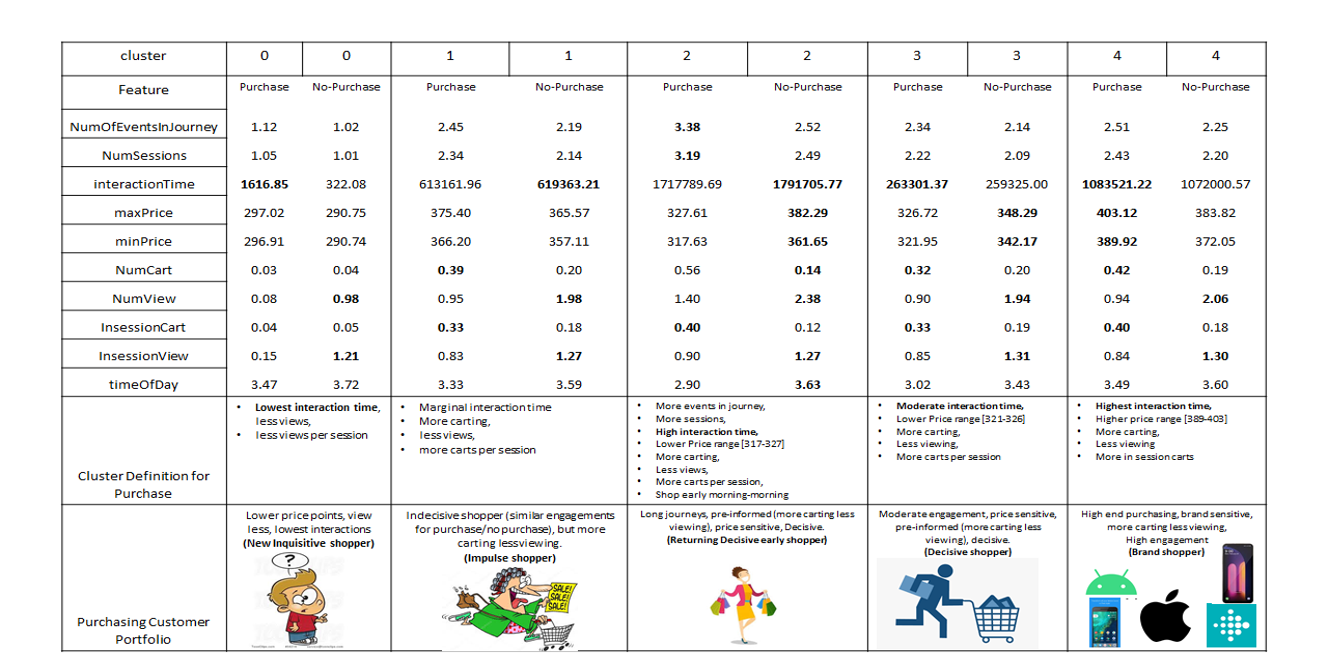}}
		\caption{User-journey feature based cluster Definitions.}\label{cosm_cluster}
       \vspace{-0.3cm}
\end{figure*}

 Next, we analyze the samples from each user-cluster and define an aggregated identity per cluster as shown in Fig. \ref{cosm_cluster}. 

Thereafter, in Table \ref{res1} we further assess the composition of each cluster in terms of the following two metrics. 1) Fraction of all samples in each cluster denoted by $Rep=\frac{n_q}{n}, \forall q=[0,..K-1]$; 2) Purchase ratio that represents the fraction of purchasing samples per cluster $PuR=\frac{\sum_i'^{n_q} y_q(i')}{n_q}$. Here, we observe that for both data sets, one major cluster represents 91-99\% of all the data samples. Also, as the cluster size decreases, the $PuR$ increases upto 3-8 times of the overall $PuR$. 
\begin{table}[ht!]
\centering
\caption{User-Cluster Definitions based on $Rep$ and $PuR$.}
\scalebox{0.99}
    {
 \begin{tabular}{|c|c|c|}
 \hline
$Cluster ID$&$Rep$&$PuR$\\ \hline
\multicolumn{3}{|c|}{Cosmetics Dataset}\\ \hline
1 (New Shopper)&91.9&11.14\\ \hline
4 (Impulsive perusing shopper)&4.83&21.01\\ \hline
0 (educated perusing shopper)&2.19&19.45\\ \hline
3 (Intentional Shopper)&1.17&22.84\\ \hline
2 (Returning budget shopper)&0.62&32.91\\ \hline
\multicolumn{3}{|c|}{Electronics Dataset}\\ \hline
0 (New shopper)&99.09&1.35\\ \hline
3 (Decisive shopper)&0.43&6.47\\ \hline
1 (Impulsive Shopper)&0.25&6.91\\ \hline
4 (Brand Shopper)&0.18&7.68\\ \hline
2 (Returning Decisive Shopper)&0.05&8.59\\ \hline
 \end{tabular}
 }
\label{res1}
\vspace{-0.3cm}
\end{table}

Ideally, targeted promotional and marketing campaigns should be based on cluster $PuR$ along with the predictability of purchase per-cluster. In Table \ref{clusterpurchase}, we analyze the predictability of a purchasing event at per-cluster using the \textit{best data model} trained in Section \ref{classify}. Here, we observe a significant variation in $recall$ and $F1$ across clusters. For example, in both data sets, the cluster corresponding to \textit{New Shoppers} (ID is 1 for cosmetics, 0 for electronics, respectively,) has the lowest $PuR$ but this cluster is the most purchase-event predictable in terms of $recall$ and $F1$. This implies that \textit{New Shoppers} require promotional incentives to return the shopping website to increase product-level interactions (number of sessions) rather that incentives like discount coupons to convert them to other clusters that may have higher $PuR$ but low purchase predictability. Also, we observe that for the cosmetics dataset, sample neighborhoods are well defined (low variations in k-NN performances) and neighborhood based purchase prediction is more robust than feature-based prediction. 
\begin{table}[ht]
\centering
\caption{User-journey Classification Performance per-Cluster. Best values in bold.}
\scalebox{0.65}
    {
 \begin{tabular}{|c|c|c|c|c||c|c|c|c|}
 \hline
\multicolumn{9}{|c|}{Cosmetics Dataset}\\ \hline
\multicolumn{5}{|c|}{Best Model}&\multicolumn{4}{||c|}{kNN Model}\\ \hline
Cluster&Accuracy&Precision&Recall&F1&Accuracy&Precision&Recall&F1\\ \hline
1, New shopper&\bf{0.9515}&\bf{0.9039}&\bf{0.6326}&\bf{0.7443}&0.8989&0.5524&0.4896&0.5192\\ \hline
4, Impulsive shopper&\bf{0.8077}&\bf{0.7784}&0.1072&0.1869&0.7616&0.4116&\bf{0.2945}&\bf{0.3433}\\ \hline
0, Educated shopper&\bf{0.8042}&\bf{0.4363}&0.0224&0.0423&0.7719&0.3646&\bf{0.2318}&\bf{0.2834}\\ \hline
3, Intentional shopper&\bf{0.7768}&\bf{0.5352}&0.0753&0.1320&0.7348&0.3907&\bf{0.2674}&\bf{0.3174}\\ \hline
2, Returning shopper&\bf{0.7050}&\bf{0.6332}&0.2307&0.3347&0.6638&0.4867&\bf{0.4166}&\bf{0.4489}\\ \hline
\multicolumn{9}{|c|}{Electronics Dataset}\\ \hline
0, New shopper&\bf{0.9999}&\bf{0.9985}&\bf{0.9963}&\bf{0.9974}&0.9998&0.9974&0.9894&0.9934\\ \hline
3, Decisive shopper&\bf{0.9867}&\bf{0.9273}&\bf{0.8717}&\bf{0.8975}&0.9773&0.9112&0.7171&0.8012\\ \hline
1, Impulsive shopper&\bf{0.9812}&\bf{0.8726}&\bf{0.8411}&\bf{0.8535}&0.96408&0.8483&0.5873&0.6911\\ \hline
4, Brand shopper&\bf{0.9708}&\bf{0.8354}&\bf{0.7988}&\bf{0.8130}&0.9461&0.7959&0.3884&0.5203\\ \hline
2, Returning shopper&\bf{0.9291}&\bf{0.6552}&\bf{0.5055}&\bf{0.5664}&0.9107&0.5145&0.1446&0.2167\\ \hline
 \end{tabular}
 }
\label{clusterpurchase}
\vspace{-0.3cm}
\end{table}

Next, we analyze the predictive nature of each cluster in terms of cluster formation, the effort required for cluster transformation and the predictive robustness for purchase events as follows.
\subsubsection{Cluster Formation}
We analyze the structure of the user-clusters in terms of the similarity between samples of the same cluster and dissimilarity with samples from other clusters using the cluster scores of $CH$ and $SS$ described in Section \ref{metrics}. We start with samples from the two clusters with maximum sample representations and continue adding samples from additional clusters to verify if the cluster scores improve or remain similar. 
In Table \ref{form}, we observe that all the 5 clusters in each data set have similar $SS$ that implies the clusters are stable in terms of distinction from samples of other clusters. Also, we observe that the $CH$ score increases as samples of additional clusters get introduced. This further strengthens the stability of the clusters.
\begin{table}[ht]
\centering
\caption{Cluster Formation using CH and SS metrics.}
 \begin{tabular}{|c|c|c|c|c|}
 \hline
Data:&\multicolumn{4}{c|}{Cosmetics}\\ \hline
Cluster IDs&[1,4]&[1,4,0]&[1,4,0,3]&[1,4,0,3,2]\\ \hline
$CH$(x$10^4$)&16.825&32.445&50.853&68.116\\ \hline
$SS$&0.8231&0.833192&0.818430&0.8157\\ \hline
Data:& \multicolumn{4}{c|}{Electronics}\\ \hline
Cluster IDs&[0,3]&[0,3,1]&[0,3,1,4]&[0,3,1,4,2]\\ \hline
$CH$(x$10^3$)&3.286&6.337&10.906&12.280\\ \hline
$SS$&0.376&0.375&0.373&0.374\\ \hline
 \end{tabular}
 \label{form}
\vspace{-0.4cm}
\end{table}

\begin{figure*}[ht!]
    \centering
	\subfigure[Cosmetics Dataset]
	{\includegraphics[width=3.1in, height=2.3in]{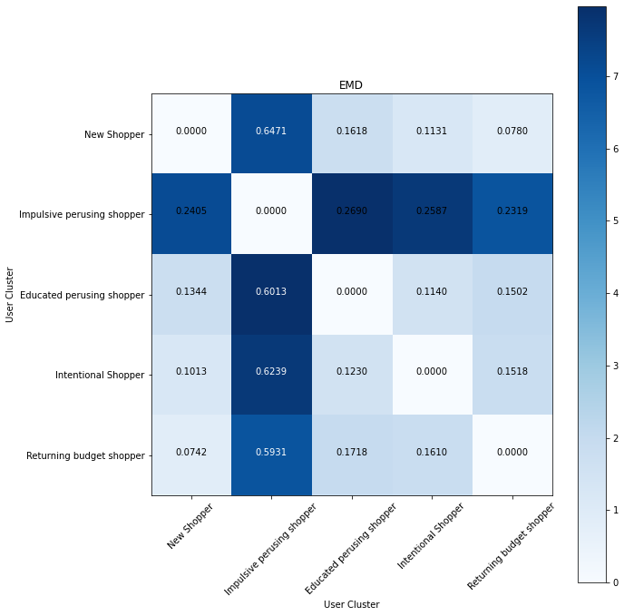}
		\label{cosmetics_emd}}
	\subfigure[Electronics Dataset]
	{\includegraphics[width=3.1in, height=2.3in]{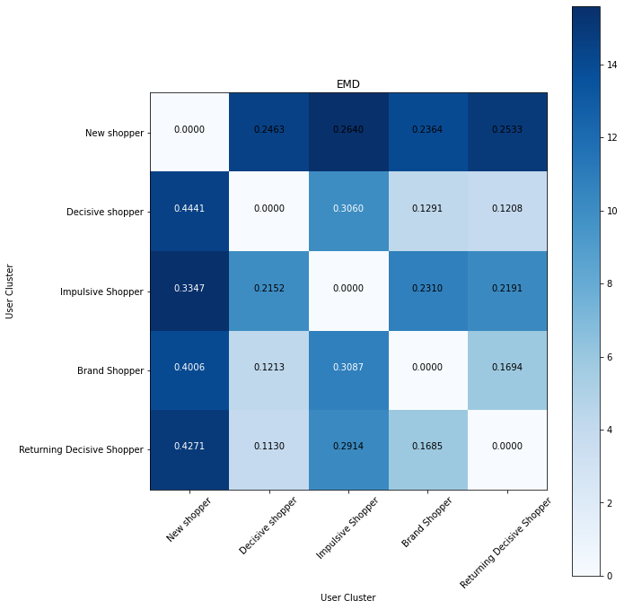}
		\label{elect_emd}}
		\caption{Normalized EMD between pairwise clusters. High values indicate difficulty to transform clusters.}\label{EMDs}
       \vspace{-0.4cm}
\end{figure*}

\subsubsection{Cluster Transformation}
So far, we have assessed static clusters for user-journeys. However, since user-journeys are constantly evolving over time, dynamically transforming customers from one cluster to another by targeted campaigns becomes a distinct possibility. Here, we analyze the effort required to transform user-journeys from one cluster to another in terms of EMD computed between each pair of user clusters in Fig. \ref{EMDs}.

We observe the following similarities across user-behaviors for both the data sets. First, \textit{Impulsive Shoppers} are most distinctive (high EMDs) from other clusters since these shoppers have greater interactions, higher views and lower carting events for non-purchasing over purchasing events. Second, all other clusters demonstrate significant variations in feature-level distribution (non-zero non diagonal EMDs), which implies varying efforts are required to transform users across specific clusters.
Additionally, from Fig. \ref{EMDs} (b) and Fig. \ref{cosm_cluster} (b), we observe that for the electronics dataset, \textit{New Shoppers} are highly distinctive at feature level, with high views for non-purchases and no other visible trend in features. This cluster requires the highest amount of \textit{nudge} to convert to other clusters. Also, we observe that \textit{Decisive Shoppers} and \textit{Brand Shoppers} have similar purchase predicting nature (low EMD between these clusters).

\subsubsection{Predictive Robustness}
\begin{figure*}[htp]
    \centering
	\subfigure[$accuracy$ for PLL on Cosmetics dataset.]
	{\includegraphics[width=3in, height=1.5in]{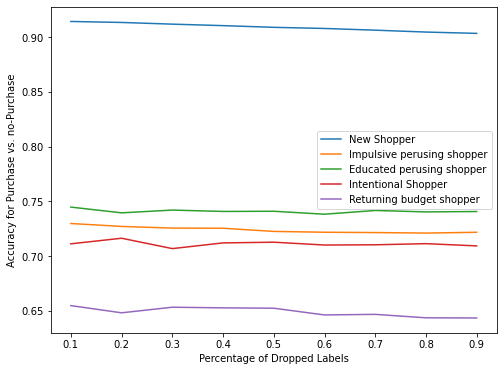}
	\label{cosmetics_plla}}
	\subfigure[$F1$ for PLL on Cosmetics dataset.]
	{\includegraphics[width=3in, height=1.5in]{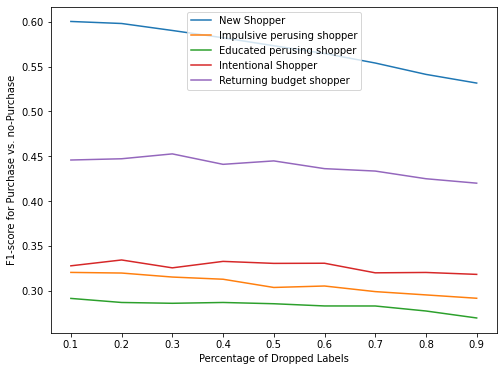}
	\label{cosmetics_pllf}}
	\subfigure[$accuracy$ for PLL on Electronics dataset.]
	{\includegraphics[width=3in, height=1.5in]{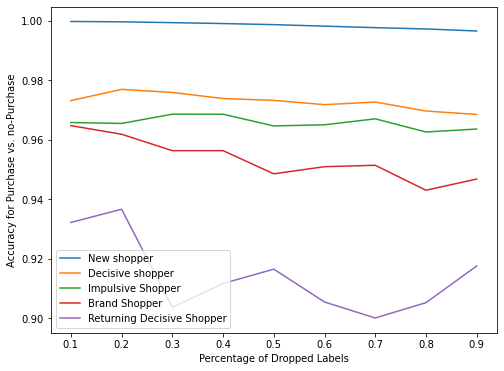}
	\label{elect_plla}}
	\subfigure[$F1$ for PLL on Electronics dataset.]
	{\includegraphics[width=3in, height=1.5in]{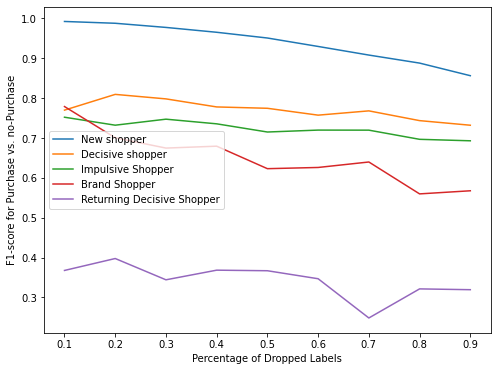}
	\label{elect_pllf}}
		\caption{PLL analysis for predictive robustness per user cluster.}\label{PLL}
      \vspace{-0.4cm}
\end{figure*}

Finally, we assess the impact of sample neighborhoods on new/unlabelled data for each user-cluster using PLL. The $accuracy$ and $F1$ plots when a fraction $p$ of randomly sampled labels are dropped and recomputed using k-NN, for $p=[0.1, 0.2..0.9]$ are shown in Fig. \ref{PLL}.
While the EMD analysis provides a snapshot of the inter-convert ability across clusters, the PLL curves provide a more dynamic view of the predictive stability for purchasing events per-cluster based on sample neighborhoods to support new/ongoing/unlabelled user-journeys.
From Fig. \ref{PLL}, we observe that \textit{New Shoppers} are the most predictable and stable cluster across datasets. However, from Fig. \ref{PLL} (a), (b), we observe that for the cosmetics data set, \textit{Returning Shopper} cluster has significantly higher $PuR$ of 32.9 over other clusters  (general $PuR$ range \{19-22\}), which leads to higher relative $F1$ scores over relative $accuracy$ for this cluster. Also, all clusters apart from the \textit{Impulsive Shopper} cluster, have relatively stable neighborhoods (slight variation in PLL curves), which further strengthens or previous finding from EMD analysis that \textit{Impulsive Shoppers} are most different from other clusters at feature and neighborhood levels.

Further, from Fig. \ref{PLL} (c), (d) we observe that for the electronics dataset, the cluster ordering in $accuracy$-based PLL curves and $F1$-based PLL curves remain the same. This is intuitive since for this dataset, the $PuR$ has a narrow variation range of \{6.4-8.5\} for non-\textit{New Shopper} clusters, which implies similar purchase to non-purchase ratios in most clusters. Further, for the electronics dataset, \textit{Brand Shopper} cluster has a highly unpredictable neighborhood, implying more product-level features may be necessary to accurately classify purchase events for this cluster. Thus, predictive models, EMD and PLL-based cluster analysis together can better provide a holistic understanding of the varying shopping behaviors. 

\section{Conclusions and Discussion}
In this work we present an analytical system for prediction of purchase vs. non-purchasing events at a session and user-journey level that can scale across product-level, shopping portal-level and data-size specifications. We analyze two data sets, cosmetics (2GB) \cite{data1} and electronics (15GB) \cite{data2}, with the following three major conclusions.

First, at session-level we find that session-level feature based LSTM models have 91-98\% $accuracy$ and 73-99\% $recall$ for predicting a purchase event compared to event-based sequence models in \cite{clicknn} that have 90-92\% $accuracy$ and 14-25\% $recall$, respectively. Also, our analysis shows that session-level purchase prediction is highly sensitive to shopping platform-related features. For instance, for the cosmetics dataset, the \textit{remove\_from\_cart} event type leads to better classification performance when compared to the electronics dataset that does not have this event type. Thus, we conclude that session-level features, and variations in event-types can influence session-based \textit{nudge models}.

Second, we analyze the user-product interaction journey using supervised and unsupervised methods, such that five distinctive clusters representing specific purchasing behaviors are identified. Next, we use the TPOT AutoML package \cite{TPOT} to fit the best classifiers for predicting a purchasing journey. Our analysis shows that for journey level predictions, product-level features such as variations in product cost, brand etc. represented in the electronics data set are significant for purchase predictions ($accuracy$/$recall$ of 99/98\%). Also, we observe that purchase prediction can vary significantly across clusters. Thus, purchase predictability per customer cluster plays a key role in designing effective strategic marketing campaigns.
 
Third, we analyze each user-behavior cluster in terms of cluster formation, capability for transformation to other clusters and predictive robustness using semi-supervised learning (PLL method). We observe that for both data sets, majority of the user-journeys samples belong to clusters representing \textit{New Shoppers}, who have a higher tendency to research a product than to make the actual purchase. Also, there are other minority clusters that demonstrate varying degrees of engagement and purchasing intent. Marketing campaigns for users must consider not only the purchase ratio per-cluster but also the capability of the cluster to handle new/unseen/unlabelled or ongoing user journeys. 
For instance, the \textit{New Shopper} clusters have least purchase to non-purchase journey ratio, but they are most stable to unlabelled data. Also, we observe that the \textit{Impulsive Shopper} cluster is significantly different from the others in terms of the EMD metric, which makes these shoppers easy to detect but difficult to convert. 
In this work, session-level and user-journey level analysis has been kept separate. Future works may be directed towards utilizing the journey-level cluster information combined with session-level features for enhanced nudge modeling at session-level.
\bibliographystyle{IEEEtran}
\bibliography{papers}

\begin{thebibliography}{10}
\providecommand{\url}[1]{#1}
\csname url@samestyle\endcsname
\providecommand{\newblock}{\relax}
\providecommand{\bibinfo}[2]{#2}
\providecommand{\BIBentrySTDinterwordspacing}{\spaceskip=0pt\relax}
\providecommand{\BIBentryALTinterwordstretchfactor}{4}
\providecommand{\BIBentryALTinterwordspacing}{\spaceskip=\fontdimen2\font plus
\BIBentryALTinterwordstretchfactor\fontdimen3\font minus
  \fontdimen4\font\relax}
\providecommand{\BIBforeignlanguage}[2]{{%
\expandafter\ifx\csname l@#1\endcsname\relax
\typeout{** WARNING: IEEEtran.bst: No hyphenation pattern has been}%
\typeout{** loaded for the language `#1'. Using the pattern for}%
\typeout{** the default language instead.}%
\else
\language=\csname l@#1\endcsname
\fi
#2}}
\providecommand{\BIBdecl}{\relax}
\BIBdecl

\bibitem{onlineref}
C.~O. Sakar, S.~O. Polat, M.~Katircioglu, and Y.~Kastro, ``Real-time prediction
  of online shoppers’ purchasing intention using multilayer perceptron and
  lstm recurrent neural networks,'' \emph{Neural Computing and Applications},
  vol.~31, no.~10, pp. 6893--6908, 2019.

\bibitem{onlineref2}
C.~J. Carmona, S.~Ram{\'\i}rez-Gallego, F.~Torres, E.~Bernal, M.~J. del Jesus,
  and S.~Garc{\'\i}a, ``Web usage mining to improve the design of an e-commerce
  website: Orolivesur. com,'' \emph{Expert Systems with Applications}, vol.~39,
  no.~12, pp. 11\,243--11\,249, 2012.

\bibitem{data1}
M.~Kechinov, ``Ecommerce events history in cosmetics shop,''
  https://www.kaggle.com/mkechinov/ecommerce-events-history-in-cosmetics-shop,
  2019.

\bibitem{data2}
M.~Kechinov:, ``Ecommerce behavior data from multi category store,''
  https://www.kaggle.com/mkechinov/ecommerce-behavior-data-from-multi-category-store,
  2019.

\bibitem{clicknn}
Z.~Wu, B.~H. Tan, R.~Duan, Y.~Liu, and R.~S. Mong~Goh, ``Neural modeling of
  buying behaviour for e-commerce from clicking patterns,'' in
  \emph{Proceedings of the 2015 International ACM Recommender Systems
  Challenge}, 2015, pp. 1--4.

\bibitem{repeatbuy}
T.~Charanasomboon and W.~Viyanon, ``A comparative study of repeat buyer
  prediction: Kaggle acquired value shopper case study,'' in \emph{Proceedings
  of the 2019 2nd International Conference on Information Science and Systems},
  2019, pp. 306--310.

\bibitem{PLL1}
N.~Xu, J.~Lv, and X.~Geng, ``Partial label learning via label enhancement,'' in
  \emph{Proceedings of the AAAI Conference on Artificial Intelligence},
  vol.~33, 2019, pp. 5557--5564.

\bibitem{PLL2}
K.~Sun, Z.~Min, and J.~Wang, ``Pp-pll: Probability propagation for partial
  label learning,'' in \emph{Joint European Conference on Machine Learning and
  Knowledge Discovery in Databases}.\hskip 1em plus 0.5em minus 0.4em\relax
  Springer, 2019, pp. 123--137.

\bibitem{TPOT}
R.~S. Olson and J.~H. Moore, ``Tpot: A tree-based pipeline optimization tool
  for automating machine learning,'' in \emph{Workshop on automatic machine
  learning}.\hskip 1em plus 0.5em minus 0.4em\relax PMLR, 2016, pp. 66--74.

\bibitem{ranking}
A.~Idris, M.~Rizwan, and A.~Khan, ``Churn prediction in telecom using random
  forest and pso based data balancing in combination with various feature
  selection strategies,'' \emph{Computers \& Electrical Engineering}, vol.~38,
  no.~6, pp. 1808--1819, 2012.

\bibitem{cluster}
Y.~Liu, Z.~Li, H.~Xiong, X.~Gao, and J.~Wu, ``Understanding of internal
  clustering validation measures,'' in \emph{2010 IEEE International Conference
  on Data Mining}.\hskip 1em plus 0.5em minus 0.4em\relax IEEE, 2010, pp.
  911--916.

\bibitem{EMD}
A.~Irpino, R.~Verde, and F.~d.~A. De~Carvalho, ``Dynamic clustering of
  histogram data based on adaptive squared wasserstein distances,''
  \emph{Expert Systems with Applications}, vol.~41, no.~7, pp. 3351--3366,
  2014.

\end{thebibliography}
\end{document}